\documentclass[manuscript,screen]{acmart}

\AtBeginDocument{%
  \providecommand\BibTeX{{%
    \normalfont B\kern-0.5em{\scshape i\kern-0.25em b}\kern-0.8em\TeX}}}

\begin{document}

\title{How Much Automation Does a Data Scientist Want?}

\author{Dakuo Wang}
\email{dakuo.wang@ibm.com}
\orcid{0000-0001-9371-9441}
\affiliation{%
  \institution{IBM Research}
}

\author{Q. Vera Liao}
\affiliation{%
  \institution{IBM Research}
}

\author{Yunfeng Zhang}
\affiliation{%
  \institution{IBM Research}
}

\author{Udayan Khurana}
\affiliation{%
  \institution{IBM Research}
}

\author{Horst Samulowitz}
\affiliation{%
  \institution{IBM Research}
}

\author{Soya Park}
\affiliation{%
  \institution{IBM Research}
}

\author{Michael Muller}
\affiliation{%
  \institution{IBM Research}
}

\author{Lisa Amini}
\affiliation{%
  \institution{IBM Research}
}

\newcommand{\autoai}{AutoML}

\renewcommand{\shortauthors}{Wang et al.}

\begin{abstract}
Data science and machine learning (DS/ML) are at the heart of the recent advancements of many Artificial Intelligence (AI) applications.
There is an active research thread in AI, \autoai, that aims to develop systems for automating end-to-end the DS/ML Lifecycle. However, 
\textbf{do DS and ML workers really want to automate their DS/ML workflow?} To answer this question, we first synthesize a \textbf{human-centered \autoai} framework with 6 \textit{User Role/Personas}, 10 \textit{Stages} and 43 \textit{Sub-Tasks}, 5 \textit{Levels of Automation}, and 5 \textit{Types of Explanation}, through reviewing research literature and marketing reports. Secondly, we use the framework to guide the design of an online survey study with 217 DS/ML workers who had varying degrees of experience, and different user roles ``matching'' to our 6 roles/personas. We found that different user personas participated in distinct stages of the lifecycle -- but not all stages. Their desired \textit{levels of automation} and \textit{types of explanation} for \autoai{} also varied significantly depending on the DS/ML stage and the user persona. Based on the survey results, we argue there is no rationale from user needs for complete automation of the end-to-end DS/ML lifecycle. We propose new next steps for user-controlled DS/ML automation.

\end{abstract}



\maketitle

\section{Introduction}
\begin{figure}
    \centering
    \includegraphics[width=0.75\columnwidth]{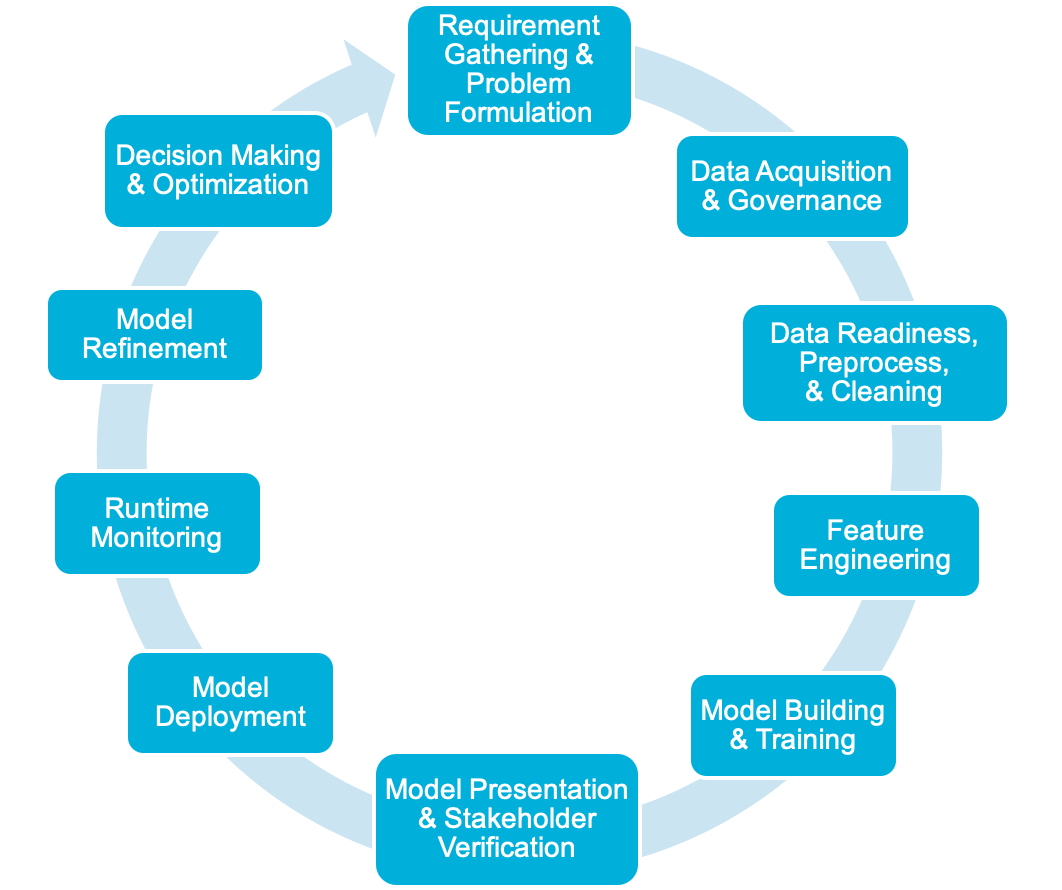}
    \caption{A 10 Stages, 43 Sub-tasks (not shown) DS/ML Lifecycle. This is a synthesized version from reviewing multiple scholarly publications and marketing reports~\cite{aggarwal2019can,wang2019humanai,8020dilemma,feurer2015efficient,berman2018realizing,gartner2020magic,lee2019human,gil2019towards}.}
    ~\label{fig:lifecycle}
\end{figure}

Data Science (DS) and Machine Learning (ML) are the backbone of today's data-driven business decision making. From a human centered perspective \cite{aragon2016developing, kogan2020mapping, muller2019human}, ML often consists of multiple stages: from gathering requirements and datasets, to
deploying a model, and to supporting human decision making---we refer to these stages together as \textbf{DS/ML Lifecycle} (as in Fig.~\ref{fig:lifecycle}). There are also diverse personas in the DS/ML team \cite{amyzhang2020} and these personas must coordinate across the lifecycle: stakeholders set requirements, data scientists define a plan, and data engineers and ML engineers support with data cleaning and model building. Later, stakeholders verify the model and domain experts use model inferences in decision making, and so on. 
Throughout the lifecycle, refinements may be performed at various stages, as needed. It is such a complex and time-consuming activity that there are not enough DS/ML professionals to fill the job demands~\cite{gartner2020magic,he2018amc}; and as much as 80\% of their time~\cite{8020dilemma} is spent on low-level activities such as tweaking data~\cite{heer2012interactive} or trying out various algorithmic options and model tuning~\cite{zoller2019survey}. These two challenges --- dearth of data scientists, and time-consuming low-level activities --- have stimulated AI researchers and system builders to explore an automated solution for DS/ML work: Automated Data Science (\textbf{\autoai})\footnote{Some researchers also refer to Automated Machine Learning (AutoML). In this paper, we use AutoDS to refer the collection of all these technologies, i.e., AutoML, AutoDS, AutoAI.}.

Several \autoai{} algorithms and systems have been built to automate the various stages of the DS/ML lifecycle~\cite{zoller2019survey}. For example, the ETL (extract/transform/load) task has been applied to the \textit{data readiness, preprocessing \& cleaning} stage, and has attracted research attention~\cite{kougka2018many}. Another heavily investigated stage is \textit{feature engineering}, for which many new techniques have been developed such as deep feature synthesis~\cite{kanter2015deep}, one button machine~\cite{lam2017one}, reinforcement learning-based exploration~\cite{khurana2016cognito}, and historical pattern learning~\cite{nargesian2017learning}. 

However, such work often targets only a single stage of the DS/ML lifecycle. For example, AutoWEKA~\cite{autoweka1,autoweka2} can automate the \textit{model building and training} stage by automatically searching for the optimal algorithm and hyperparameter settings, but it offers no support for examining the training data quality, which is a critical step before the training starts.

In recent years, a growing number of companies and research organizations have started to invest in driving automation across the full \textbf{end-to-end \autoai} system. For example, Google released their version of AutoML in 2018~\cite{web:googleautoml}. Startups like H2O~\cite{web:h2o} and Data Robot~\cite{datarobot} both introduced products. There are also Auto-sklearn~\cite{feurer2015efficient} and TPOT~\cite{web:tpot,olson2016tpot} from the open source community. Most of these systems aim to support end-to-end DS/ML automation. However, current capabilities are focused on the model building and data analysis stages, while little automation is offered for the human-labor-intensive and time-consuming \textit{data preparation} or \textit{model runtime monitoring} stages. 

Moreover, these works currently lack an analysis from the users' perspective: Who are the potential users of envisioned full automation functionalities? Are they satisfied with the \autoai's performance, if they have used it? Can they get what they want and trust the resulting models? These questions motivate us to pursue a systematic examination of the end-to-end \autoai{} lifecycle, but from a human-in-the-loop perspective.

In answering these questions, we first filled in the research gap by proposing a comprehensive human-in-the-loop \autoai{} framework. Secondly, we used the proposed framework to develop an online survey to gather perspectives from data science and machine learning practitioners. 

The proposed framework is synthesized from a collective effort of research publications and marketing analyses~\cite{aggarwal2019can,wang2019humanai,8020dilemma,feurer2015efficient,berman2018realizing,gartner2020magic,lee2019human,gil2019towards}. We have four dimensions: 10 lifecycle stages with 43 sub-tasks, 6 practitioner roles, 5 levels of automation (L0---L4), and 5 types of explanation.

The online survey study was conducted within a multinational IT company with a prominent focus on ML solutions. Overall, 239 respondents answered the survey during a 7-day deployment. These respondents are professionals from diverse backgrounds, such as DS experts, ML engineers, and designers. The only qualifying criteria is that each of them has previously worked on DS/ML projects. The results support the completeness of our framework, as it covers all the respondents' personas, the stages they participated in, and the tasks they performed. Further analyses of the survey results suggest that the desired levels of automation and types of explanation for \autoai{}  vary significantly, depending on which stages of lifecycle the respondent performs and their role. Thus, we argue full automation (L4) of the entire end-to-end DS/ML lifecycle is not currently a priority for data science and machine learning practitioners; and a fully automated \autoai{} with no human involvement (L4) should not be the priority of \autoai{} researchers and system builders. 

At the end of this paper, we leverage this framework to shed light on future research directions for Human-in-the-Loop ML researchers and HCI researchers to consider.

\section{Related Work}
In this paper, we will first review the related HCI and CSCW literature around our research topic. Because AutoML is a recent topic, we only found a couple of CHI and CSCW papers directly relevant to it. Hence, we expand our literature review to include broader topics such as human-in-the-loop machine learning and AI,  automation systems, and human-AI interactions. In the following sub-section, we briefly review the technical advancement of \autoai{} techniques so that our readers can have a general common ground understanding about the core technology. The third category of literature is on explainability of intelligent systems, especially recent work on explainable AI (XAI). Users need to \textit{understand} an automated system in order to trust it and act on it \cite{JaimieDrozdal2020}, so we consider explainability as an essential piece of the puzzle in human-in-the-loop \autoai{} research.


\subsection{Data Science Team and Data Science Lifecycle}
Data science and machine learning are complex practices that requires a team with interdisciplinary background~\cite{muller2019datascience,muller2019human, green2015mining, clarke2019training,saltz2018exploring} and skills~\cite{passi2017data,passi2018trust, berrar2019incorporating}. For example, the team often include stakeholders who have deep domain knowledge and own the problem~\cite{gartner2020magic, green2015mining}; it also must have DS/ML professionals who can actively work with data and write code~\cite{sutton2018data,kery2019towards}.

Due to the interdisciplinary and complex nature of the DS/ML work, teams need to closely collaborate across different job roles, and the success of such collaboration directly impacts the DS/ML project's final output model performance~\cite{passi2017data,hou2017hacking,mao2019}. For example, Passi and Jackson showed the importance of including diverse expertise in many complex data science projects~\cite{passi2018trust}. Hou et al. studied how citizen data scientists, non-profit organization (NPO) managers as stakeholders, and hackathon organizers work together in DS projects~\cite{hou2017hacking}. They reported that DS and NPO managers' collaboration can benefit from having hackathon organizers acting as ``brokers'' in this interdisciplinary collaboration territory~\cite{wenger2010communities}  (see ~\cite{paepcke1996information} for a review of the broker theory
). Mao et al.~\cite{mao2019} looked into the collaboration between data scientists and bio-medical scientists (as stakeholders) in data-driven scientific research projects, and they found that it is nearly impossible to build a \textit{static} common ground for these two groups, but instead, a \textit{fluid and always-evolving} common ground is needed throughout these projects. Borgman et al.~\cite{borgman2012s} studied who provides what kinds of data in data science teams, and argued for a combination of data science knowledge with domain knowledge. These studies exemplify that DS projects are multi-disciplinary collaborations that require considerable coordination and communication~\cite{viaene2013data,stein2017make}. Thus, in this paper, we use ``\textbf{data scientist}'' to refer to the people who code and build models, and we use ``\textbf{DS/ML practitioners}'' (ref.~\cite{muller2019datascience}) to collectively refer to all the people who participate in a project, including various \textbf{roles/personas}: data scientists, engineers, designers, and domain experts. 

DS/ML projects are not only interdisciplinary, but are also highly iterative staged processes~\cite{amyzhang2020,muller2019datascience,rule2018exploration,wang2019data,wang2019humanai,feinberg2017design,kiss2013evaluation}. Existing literature coarsely divides the DS/ML workflow into stages about data, about model, or about the domain context~\cite{feinberg2017design,muller2019datascience,guo2011proactive}. For example, Feinberg described DS as the \emph{design} of data~\cite{feinberg2017design}, and Muller extended this claim and articulated the DS process as five human-centric steps: discovery of data, capture of data, design of data, curation of data, and even creation of data~\cite{muller2019datascience}. 

Other researchers have defined finer-grained stages for the tasks that the DS team performs~\cite{song2016big,van2014data,amyzhang2020,grappiolo2019semantic,kiss2013evaluation}. For example, Zhang et al. defined three phases (\textit{preparation, modeling, and deployment}) and ten steps (e.g., \textit{data cleaning, feature engineering, model selection, and runtime monitoring}) in the DS workflow ~\cite{amyzhang2020}. Patel et al. specifically explored the human-constructive activities during \textit{feature engineering}~\cite{patel2008investigating}. Song et al. and Van der Aalst focused on model selection after the data is ready and in modeling stage~\cite{song2016big,van2014data}, and d'Alessandro et al. and Walker emphasized on testing for bias in the development stage of model~\cite{d2017conscientious,walker2015professionalisation}.

Researchers refer to this staged process of DS/ML as ``workflow''~\cite{amyzhang2020}, ``lifecycle''~\cite{arnold2020towards}, ``steps''~\cite{patel2008investigating}, ``process''~\cite{wang2019atmseer}, etc. In this paper, we use the term ``\textbf{DS/ML Lifecycle}'' to collectively refer to the entire flow of a DS project. Within the lifecyle, we use ``\textbf{Stage}'' to describe the conceptual separation of tasks, and use ``\textbf{Sub-Tasks}'' to describe the detailed action or task that DS/ML practitioners performed in it. It is obvious to speculate that the DS/ML practitioner's role/persona and lifecycle stage may have an interaction effect that can further complicate the collaboration, refer to~\cite{amyzhang2020} for a full review. 

In addition to these empirical explorations, many HCI researchers have also started to build tools and platforms to support DS/ML practitioners ~\cite{grappiolo2019semantic,wang2019data,wang2020callisto,chang18}. Jupyter Notebook~\cite{web:jupyter} and its customized version Google CoLab~\cite{web:colab} JupyterLab~\cite{web:jupyterlab} are widely adopted by data scientists and ML engineers. As opposed to the traditional console-oriented programming environment (e.g., Eclipse), these tools provide a richer user interface that allows coders to write code in coding cells, track results in output cells, and document a narrative in markdown cells~\cite{kross2019practitioners}. Despite its wide adoption, Notebooks still fall short in some areas. For example, many data scientists neglect to fully utilize the markdown cells to write the narrative~\cite{rule2018exploration}, and some domain experts do not have a friendly experience with Notebooks~\cite{grappiolo2019semantic}. Researchers are investigating solutions to these issues. For example, to support better DS team collaboration, Wang et al. built two experimental features: one is to enable multiple data scientists to edit the same Notebook synchronously~\cite{wang2019data}; the other one is to allow Notebooks to capture the conversations happening between multiple data scientists and translate them into documentations in the Notebook~\cite{wang2020callisto}. Both features have proven useful in followup user evaluations. 

In this paper, we synthesize knowledge and findings from existing literature on Human Centered Data Science~\cite{aragon2016developing,kogan2020mapping,muller2019human} and Human Centered Machine Learning ~\cite{sacha2016human} to develop a customized framework and taxonomy for human-in-the-loop \autoai{} research in Section~\ref{framework}.

\subsection{Automated Data Science}

In this sub-section, we conduct a brief literature review of recent \autoai{} research for the HCI and CSCW audience. As shown in Fig.~\ref{fig:lifecycle}, DS/ML is an iterative and staged process. It often starts with the stage of requirement gathering \& problem formulation, followed by data cleaning and engineering, model training and selection, model tuning and ensembles, and finally deployment and monitoring.  Automated Data Science (\autoai) is the endeavor of automating each stage of this process separately or jointly. 

The {\em Data cleaning} stage focuses on improving data quality. It involves an array of tasks such as missing value imputation, duplicate removal, noise correction, invalid values and other data collection errors. AlphaClean~\cite{krishnan2019alphaclean} and HoloClean~\cite{holoclean} provide representative examples of automated data cleaning. The \textit{Data fusion} stage deals with combining various data sources, and example automations includes  DSM~\cite{kanter2015deep} and OneBM~\cite{lam2017one}. 
The {\em Feature engineering} stage is a complicated and time consuming task~\cite{kaggle2018survey, khurana2018transformation} which involves altering the feature space to improve modeling accuracy. Automation has been achieved through approaches like reinforcement learning~\cite{khurana2016cognito, khurana2018feature}, trial and error methodology~\cite{explorekit}, historical pattern learning~\cite{lfe} and more recently through knowledge graphs~\cite{kafe}. The \textit{Hyperparameter selection} stage is used to fine tune a model or the sequence of steps in a model pipeline. Several automation strategies have been proposed, including grid search~\cite{hesterman2010maximum}, random search~\cite{bergstra2012random}, evolutionary algorithms~\cite{escalante2009particle, claesen2014easy}, and finally sequential model based optimization methods~\cite{hutter2011sequential, bergstra2011algorithms}. 
Sabharwal et al.~\cite{sabharwal2016selecting} present an automated {\em algorithm selection} method for efficiently finding the best estimator without completely evaluating all candidates on full data.

\autoai{} has witnessed considerable progress in recent years, in research as well as application in commercial products. 
Various \autoai{} research efforts have moved beyond the automation on one specific step. Joint optimization~\cite{liu2019admm}, a type of Bayesian-Optimization-based algorithms, enables \autoai{} to automate multiple tasks together. For example, AutoWEKA~\cite{autoweka1,autoweka2}, Auto-sklearn~\cite{feurer2015efficient}, and TPOT~\cite{web:tpot,olson2016tpot} all automate the \emph{model selection}, \emph{hyperparameter optimization}, and \emph{ensembling} steps of the data science pipeline. The result coming out of such \autoai{} system is called a ``\textbf{model pipeline}''. A model pipeline is not only about the model algorithm; it emphasizes the various data manipulation actions (e.g., filling in missing value) before the model algorithm is selected, and the multiple model improvement actions (e.g., optimize the best values for model's hyperparameters) after the model algorithm is selected.

Amongst these advanced \autoai{} systems, Auto-sklearn~\cite{automl} and Auto-WEKA~\cite{autoweka1,autoweka2} are two open source efforts. Both use the sequential-parameter-optimization algorithm~\cite{automl,autoweka2}. This optimization approach generates model pipelines by selecting a combination of model algorithms, data pre-processors, and feature transformers. Their system architectures are both based on the same general-purpose-algorithm-configuration framework, SMAC (Sequential Model-based Optimization for General Algorithm Configuration)~\cite{smac}. In applying SMAC, Auto-sklearn and Auto-WEKA translate the model selection problem into a configuration problem, where the selection of the algorithm itself is modeled as a configuration.

Auto-sklearn supports warm-starting the configuration search by trying to generalize configuration settings across data sets based on historic performance information. Fusi et al.~\cite{recommendersystem2018} leverage historical information to build a recommender system that can navigate the historical information more efficiently. This approach is effective in determining a pipeline, but is also limited because it can only select from a pre-defined and limited set of pre-existing pipelines.

To enable \autoai{} to dynamically generate pipelines instead of only to select pre-existing pipelines, Droi et al. ~\cite{alphadgil20193m} were inspired by AlphaGo Zero~\cite{alphazero} and its pipeline generation algorithm. modelded as a single-player game. So the pipeline is built iteratively by selecting a set of actions (insertion, deletion, replacement) and a set of pipeline components (e.g., logarithmic transformation of a specific predictor or ''feature''). Khurana et al.~\cite{khurana2018ensembles} extend this idea to use a reinforcement learning approach, so that their final pipeline outcome is an ensemble of multiple sub-optimal pipelines, but that final pipeline has a state-of-the-art model performance when compared to other approaches.

Model ensembles have become a mainstay in ML with all recent Kaggle competition winning teams relying on them. Many \autoai{} systems generate a final output model pipeline as an ensemble of multiple model algorithms instead of a single algorithm. More specifically, the ensemble algorithm includes: 1) \textit{ensemble selection}~\cite{ensembleselection}, which is a greedy-search-based algorithm that starts with an empty set of models, incrementally adds a model to the working set, and selects that model if such addition results in improving the predictive performance of the ensemble; 2) and, \textit{genetic programming algorithm} (e.g., as used in TPot~\cite{olson2016tpot}), which does not create an ensemble of multiple model algorithms, but it can compose derived model algorithms. An advanced version of the genetic programming algorithm is presented in~\cite{geneticensembles} that uses multi-objective genetic programming to evolve a set of accurate and diverse models via introducing bias into the fitness function accordingly.

For a more comprehensive survey of \autoai{} techniques, we refer the reader to Zoller et al.~\cite{zoller2019survey}.

\subsection{Human-in-the-Loop Explainable \autoai}
With the recent promising advancement of \autoai{} research, more and more researchers have started to explore the possibility of a full end-to-end \autoai{} system. In that vision, from the \textit{requirement gathering and problem formulation}, to \textit{data cleaning}, to \textit{model building} and \textit{deployment}, and eventually to \textit{decision making}, no human is needed in this process~\cite{aggarwal2019can,arnold2020towards}. Some companies have also expressed their interest in \autoai{} systems that can fully autopilot the end-to-end DS/ML lifecycle~\cite{niemer_chen_naqvi_kaynat_zhu_2019}.

We argue that a fully automated end-to-end DS/ML lifecyle may not be what DS/ML practitioners want in practice. Even for traditional AI/ML practices, users reported difficulties in understanding AI/ML systems functionality~\cite{kaur2020interpreting}, and they found it difficult to trust a ML model or an AI system that they do not understand~\cite{shneiderman2020human,miller2019explanation}. That is why a group of AI and HCI researchers started working on the human-in-the-loop (HITL) AI/ML research thread in recent years~\cite{gil2019towards,amershi2019guidelines,wang2019atmseer,zoller2019survey,cai2019human,wang2019data,karl2020,wang2020human}. For example, Gil et al.  proposed design guidelines for developing human-guided ML systems based on their own experience and on surveying the research literature~\cite{gil2019towards}; and Amershi et al. proposed AI design guidelines which emphasized the human labelers' and coders' interactions with the system ~\cite{amershi2019guidelines}. 

Following this trend, in this paper we study the \textbf{Human-in-the-Loop \autoai} research topic. We believe an end-to-end automated DS/ML lifecycle will benefit from human DS/ML practitioners in the loop. The aforementioned HITL AI/ML literature provided inspirational but limited knowledge for us to understand this new research topic, because: 
\begin{enumerate}

\item The target user population is different. the HITL-AI design guidelines~\cite{amershi2019guidelines} emphasized the design of applications for end users, such as doctors and customers, to help them understand the AI recommendation and to make a better decision. The HITL-ML literature~\cite{holzinger2016interactive} focuses on building interactive user interfaces either to support data labelers to efficiently label data, or to support ML engineers to check model performance via a visualization. However, in the end-to-end \autoai{} research, the target users include both traditional ML engineers and data labelers, but also other DS works such as salespeople, citizen data scientists, or business stakeholders. They have very different expectations and requirements, and sometimes their interests may conflict with each other~\cite{mao2019}. How can we design an \autoai{} system for every user role that is involved in a DS/ML project, if the ultimate goal is to support the automation of the end-to-end DS/ML lifecyle?

\item The level of complexity and level of automation are different. In the traditional ML context, people provide one input data point, and it generates one prediction outcome. Thus people can use this relational projection to rationalize how the model works. In fact, this is the idea behind the t-SNE model interpretation method~\cite{lipton2018mythos}. But, in an \autoai{} workflow, the ML model is simply a component of the \autoai's output pipeline. Interpreting and controlling one ML model is hard, to interpret and to control an \autoai{} process that simultaneously can generate hundreds of ML modes is harder.

\item Last but not least, how will DS/ML practitioners perceive and interact with such a fully end-to-end \autoai{} system? The autopilot level of intelligence may dramatically change how these DS/ML practitioners do their job, and may even threaten their job security in the long term. On the other hand, an autopilot \autoai{} may help today's non-technical DS/ML practitioners, such as stakeholders, by reducing the boundary for them to build a model on their own. But, foremost we need to answer the fundamental research question: \textbf{Do data science and machine learning workers really want \autoai{} to automate the end-to-end lifecycle?}
\end{enumerate}

To the best of our knowledge, there is no existing literature discussing the \textbf{End-to-End \autoai} from a user perspective. There are two notable recent works started the preliminary investigation along this direction, but each has its own limitations:

~\citet{wang2019humanai} used a think-aloud method and interviewed 30 professional DSs about their perceptions on a \autoai{} system demo. While these DSs reported in general that they welcomed the \autoai's support in their DS/ML projects, none of them had actually used \autoai{} before. That inexperience may have contributed to the result that they do not think they would trust the model results built by automation as much as they trust models built by other human data scientists. Our work in this paper leverages on an online survey research method to solicit user needs and perceptions of the end-to-end \autoai{} with a broad and diverse set of user personas, and reports their current experiences and future preferences of such automation tools.

Lee et al. aimed to propose a design guideline for \autoai{} systems with HITL features ~\cite{lee2019human}. To do so, the authors introduced three levels of automation (User-Driven, Cruise-Control, and Autopilot), but they did not explain the theoretical origin of this classification. Also, the innovation of the framework stopped at the invention of automation level, but fall short on covering other important dimensions such as: who are the users? and how does the automation level interplay with different user roles? Furthermore, the discussion regarding transparency and interpretability did not provide actionable guidelines for designers and engineers to move forward. Thus, this paper reads more like a summary of the various \autoai{} techniques the authors have built, and the framework is designed to pack all these works in. Our work builds on top of the level of automation dimension discussed in this work, as well as referencing other works on levels of automation~\cite{mackeprang2019discovering,parasuraman2000model} mixed-initiative~\cite{horvitz1999principles}, agency and direct manipulation~\cite{shneiderman1997direct}, and borrows concepts from the levels of autonomous cars~\cite{young2007driving}. In addition, our framework also extends to cover dimensions of user roles, DS/ML lifecycle, the various sub-tasks that users perform in the lifecycle.

Our work explored people's information needs for explanation with human-in-the-loop \autoai. For autonomous technology that could act on its own, people ought to be able to \textit{understand} how the technology works and might affect them, in order to trust it, better work with it, and ultimately feel in control of the outcome that they are accountable for~\cite{abdul2018trends,bellotti2001intelligibility,liao2020questioning}. The HCI community has a long history of studying and pushing for explainable, intelligible and accountable technologies, from early-generation expert systems~\cite{swartout1983xplain}, context-aware technologies~\cite{lim2009and,lim2009assessing}, recommender systems~\cite{tintarev2007survey,pu2007trust} to the current focus on explainable or interpretable AI, as detailed in a survey paper by Abdul et al.~\cite{abdul2018trends}. Recently, the wide adoption of increasingly capable, but complex and opaque AI technologies, such as those using deep neural networks, has spurred a surge of interest in explainable AI (XAI) systems.  Started as a sub-field of AI, XAI work aims to produce techniques and methods that make AI's decisions and behaviors understandable by humans. A review of the technical field of XAI can be found in recent survey papers~\cite{guidotti2018survey,adadi2018peeking,arya2019one,gilpin2018explaining,carvalho2019machine}. However, many have criticized the disconnect of current XAI field with user needs, calling for interdisciplinary collaboration between the communities of AI, HCI and the social sciences ~\cite{miller2019explanation,doshi2017towards}, as well as user-centered approaches~\cite{liao2020questioning,wang2019designing}. 

Importantly, explainability does not equate to complete transparency of every aspect of an autonomous system. Studying explanatory debugging systems to enable users to debug ML system through its explanations, Kulesza et al.~\cite{kulesza2015principles} defined a ''good'' explanation as
truthful and complete, but not overwhelming. Hoffman~\cite{hoffman2018metrics} emphasizes that the effectiveness of an explanation depends on the user and context, described as “triggers” by  representing “tacitly an expression of a need for a certain kind of
explanation...to satisfy certain user purposes of user goals.” In HCI works that studied user needs for explainable systems, there is a tradition of using \textit{prototypical questions} to represent different types of needs. For example, in the foundational work by Lim and Dey~\cite{lim2009and,lim2009assessing} on intelligible context-aware applications, user needs for different types of intelligibility were represented as \textit{what input/output}, \textit{how} \textit{why}, \textit{why not}, \textit{what-if},\textit{what else}, \textit{certainty} and \textit{control} questions. Rader et al.~\cite{rader2018explanations} used a similar taxonomy of question types to understand social media users' needs for explanation to support algorithmic transparency. Liao et al.~\cite{liao2020questioning} adapted Lim and Dey's taxonomy to understand how explainability needs are presented and vary in different AI products. 

\section{A Comprehensive \autoai{} Theoretical Framework: From a User-Centered Perspective} \label{framework}

The first goal of this paper is to propose a comprehensive human-in-the-loop \autoai{} framework to structure the knowledge and to guide the design considerations for building an end-to-end \autoai{} system. As aforementioned, a couple of recent works provide valuable but incomplete insights toward this direction. ~\citet{wang2019humanai} learned that users prefer different levels of automation at the 4 different stages of the DS/ML Lifecycle model; Lee et al. proposed a 5-stages-10-subtasks and 3 levels of automation model ~\cite{lee2019human}; Gil et al.  proposed a 3-stages-9-sub-tasks model (data use, model development, model interpretation) lifecycle interplay with user's rationale and intended guidance dimensions ~\cite{gil2019towards}. 
We agree with these works, that a framework should include both DS/ML lifecycle stages and levels of automation, and we want to further complete these definitions. In addition, multiple HCI papers have reported that DS/ML is a ``team sport'' that requires multiple user roles with a diverse background~\cite{muller2019datascience,mao2019,patil2011building,kery2018story,rule2018exploration,amyzhang2020}. Similarly, various marketing reports and business analyses suggest to categorize DS/ML team members into different personas (e.g., data scientist or domain experts) ~\cite{gartner2020magic}. Thus, we argue that the user role should also be a dimension in the framework. Our argument is echoed by a several \autoai{} research reports, which claim to ``democratize AI'' by enabling non-technical users to build ML with \autoai~\cite{lam2017one, mackeprang2019discovering}. 

\begin{table}
\caption{A Comprehensive Theoretical Framework for Human-in-the-Loop \autoai}
\resizebox{\columnwidth}{!}{%
\begin{tabular}{|l|l}
\hline
\textbf{Level of Automation (5 Levels)}                                 & \multicolumn{1}{l|}{\textbf{Lifecycle (10 Stages, 43 Sub-Tasks)}}  \\ \hline
L0 No automation                                                        & \multicolumn{1}{l|}{Requirement Gathering and Problem Formulation} \\ \hline
L1 Human-directed automation                                            & \multicolumn{1}{l|}{Data Acquisition and Governance}               \\ \hline
L2 System-suggested Human-directed automation                           & \multicolumn{1}{l|}{Data Readiness and Preparation}                \\ \hline
L3 System-directed automation                                           & \multicolumn{1}{l|}{Feature Engineering}                           \\ \hline
L4 Full automation                                                      & \multicolumn{1}{l|}{Model Building and Training}                   \\ \hline
                                                                        & \multicolumn{1}{l|}{Model Presentation and Verification}           \\ \hline
\textbf{Type of Explanation (5 Types)}                                 & \multicolumn{1}{l|}{Model Deployment}                              \\ \hline
What: What kind of automatic {[}stage name{]} can the system do?        & \multicolumn{1}{l|}{Model Runtime Monitoring}                      \\ \hline
How: How does the system perform automatic {[}stage name{]}?            & \multicolumn{1}{l|}{Post-Deployment Model Refinement}              \\ \hline
Why: Why did the system execute this way and give this output?          & \multicolumn{1}{l|}{Decision Making and Optimization}              \\ \hline
What-if: Would the system execute differently if...(something changes)? &                                                                    \\ \cline{1-1}
Confidence: How confident is the system about this choice of execution? &                                                                    \\ \cline{1-1}
                                                                        &                                                                    \\ \cline{1-1}
\textbf{Personas / Roles (6 Roles)}                                     &                                                                    \\ \cline{1-1}
Expert Data Scientists / Researcher                                     &                                                                    \\ \cline{1-1}
Citizen Data Scientists                                                 &                                                                    \\ \cline{1-1}
AI-Ops / ML-Ops                                                         &                                                                    \\ \cline{1-1}
Lead / Manager                                                          &                                                                    \\ \cline{1-1}
Domain Expert / Subject Matter Expert                                   &                                                                    \\ \cline{1-1}
Non-Technical Supporting Roles                                          &                                                                    \\ \cline{1-1}
\end{tabular}%
}
\label{tab:framework}
\end{table}

As shown in Table~\ref{tab:framework}, our framework has four dimensions\footnote{For a more comprehensive list, such as the definitions of Level of Automation, Types of Explanation, the User Roles/Personas, and the Stages plus 43 Sub-Tasks, please refer to Appendix~\ref{appendix:role}}. We include the \textbf{Level of Automation} dimension. However, contrary to the three levels of Lee et al. (user-directed, cruise-control, autopilot) ~\cite{lee2019human}, we propose five levels. Our five-level definition was developed on top of Levels of Driving Automation (6 levels)~\cite{young2007driving}, and the foundational work about people interacting with general automation systems (10 levels)~\cite{parasuraman2000model}. We also borrow the principle of \textit{mixed-initiative}~\cite{horvitz1999principles} that argues for a close-coupling interaction work between the human users and the system. For example, our Level 1 to Level 3 are human-directed automation, system-suggested human-directed automation, and system-directed automation, with each level shifting a bit more control from human to automation. However, the mixed-initiative frameworks (\cite{horvitz1999principles,lee2019human}) do not distinguish these three levels and combine them into one cruise-control automation. Also, comparing to the 10 levels of general automation and the 6 levels of autonomous driving, our 5 levels are articulated for the AutoML domain, and eliminate the redundant and irrelevant scales. Thus, this framework can guide actionable design implications for \autoai. For more detail, refer to Appendix~\ref{appendix:level}.

For the role dimension in Table~\ref{tab:framework}, we referred to both HCI literature (e.g., ~\cite{muller2019datascience,mao2019,rule2018exploration}) and AI literature (e.g., ~\cite{khurana2018feature,gil2019towards}), as well as market reports (e.g., Gartner report~\cite{gartner2020magic} and Kaggle Survey~\cite{kaggle2018survey}). The listed 6 roles together with their definitions (in Appendix~\ref{appendix:role}) are sufficient to cover all the DS/ML team member roles described in those sources. 

For the lifecycle stages and sub-tasks dimension, we again referred to the aforementioned HCI and AI literature and various market reports. We also conducted expert interviews with 8 AutoML experts in the field. Consolidating all the literature and expert feedback, we developed this lifecycle dimension with 10 stages 43 sub-tasks. We believe the current result should cover most of the activities in a DS/ML project. Here we list the 10 stages and their examples, with more details in Appendix ~\ref{appendix:lifecycle}.

\begin{enumerate}
    \item Requirement Gathering and Problem Formulation (e.g, System can gather domain knowledge or regulation requirements and define a data science problem.)
    \item Data Acquisition and Governance (e.g, System can find relevant dataset and combine datasets for given a data science problem.)
    \item Data Readiness, Preparation, and Cleaning (e.g, System can assess data quality, detect data noice, and clean the data. ) 
    \item Feature Engineering (e.g, System can apply various feature transformation and encoding functions, and select the best subset of features.)
    \item Model Building and Training (e.g, System can split train-test dataset, select the best algorithm or neural architecture, and find an optimized set of hyperparameters).
    \item Model Presentation and Verification (e.g, System can prepare reports and dashboard for a model, and verify model with stakeholders.)
    \item Model Deployment (e.g, System can dockerize model as microservice, deploy, and conduct A/B test with different versions of the model.)
    \item Model Runtime Monitoring (e.g, System can monitor performance drift, security, and fraud risk.)
    \item Post-Deployment Model Refinement (e.g, System can refine the model with deployment data to optimize Key-Performance-Indicator-based (KPI) not only for accuracy metrics. )
    \item Decision Making and Optimization (e.g, System can automatically collect domain knowledge for contextualize decision making and further support decision optimization. )
\end{enumerate}

\section{Method}
217 people in data science roles responded to our survey. 

\subsection{Survey Design}

\label{survey_design}
The survey consisted of two sections: 1) It started with two questions asking participants' roles and experiences in the company and in their DS/ML projects. If participants did not have any experience in DS/ML projects, they would be redirected to a disqualification page. For the qualified participants, then 2) the survey asked them to select which stages of the DS/ML lifecyle they had participated in the past. 

For each selected DS/ML stage, participants were asked to complete another section of the survey. This section asked them, for each of the entire stage and its sub-tasks, what level of automation they currently have in a DS/ML project; what level of automation they would prefer to have in the future; and what types of information they would need to evaluate and trust AutoML in that stage.
Specifically, the two questions were in the following form:
\begin{itemize}
    \item \textit{For \textit{[replace with stage (e.g.,Feature Engineering)]}, what level of automation do you currently use in your Data Science/ML projects?}
    \item \textit{For \textit{[replace with stage (e.g.,Feature Engineering)]}, what level of automation would you prefer having in the future?}
\end{itemize}

In total, there were 10 stages as described in Section~\ref{framework}. The participant could select and answer questions in multiple stages, and we always presented the definitiona of the five levels of automation (in Appendix~\ref{appendix:level}) at the top of each survey section for each stage.

Following the automation questions for each stage, a question inquiring about participants' need for explainability was asked. The participants answered the question by rating the importance of being provided information for the following five types of questions if the stage were to be automated: 
\begin{itemize}
    \item \textit{What}: What kind of automatic [stage name] can the system do?
    \item \textit{How}: How does the system perform automatic [stage name]? 
    \item \textit{Why}: Why did the system execute this way and provide this output?  
    \item \textit{What-if}: Would the system execute differently if...(something changes in the ML task or data)?
    \item \textit{Confidence}: How confident is the system about this choice of execution? 
\end{itemize}

All ratings were given on a 5-point Likert scale from ``Not at all important'' to ``Extremely important''. The type of questions were selected from the taxonomy of intelligibility types in Lim et al.~\cite{lim2009and,lim2009assessing} and follow-up works~\cite{liao2020questioning,rader2018explanations}, and were adapted to the context of automating DS/ML projects. For three out of the ten stages with a higher number of sub-tasks (Data Readiness and Preparation, Feature Engineering, Model Building and Training), we asked the explainability question for each sub-task. For the remaining stages, we only asked the explainability question regarding the entire stage to limit the length of the survey.

\subsection{Participants}
Participants were a self-selected convenience sample of employees in ITCorp\footnote{A multinational IT company. Pseudo name for blind review.} who subscribed to Slack channels about data science (e.g., channel-names such as ``deeplearning'', ``data-science-at-ITCorp'', ``ITCorp-nlp'', and similar, in total 5 channels). In addition, we also use snowball sampling~\cite{biernacki1981snowball} to recruit our connections and ask them to reach out to their connections. Participants worked in diverse roles and in various departments of the organization (e.g., research, engineering, health sciences, and related line-of-business). The only criteria for participant selection is that they should have participated in DS/ML projects in the past. A DS/ML practitioner is broadly defined - e.g., it could be a designer who has contributed design sketches for an AI application. Of the 239 initial respondents, 22 did not qualify for our analysis, and were excluded from further analysis.

\begin{figure}
  \centering
  \includegraphics[width=\linewidth]{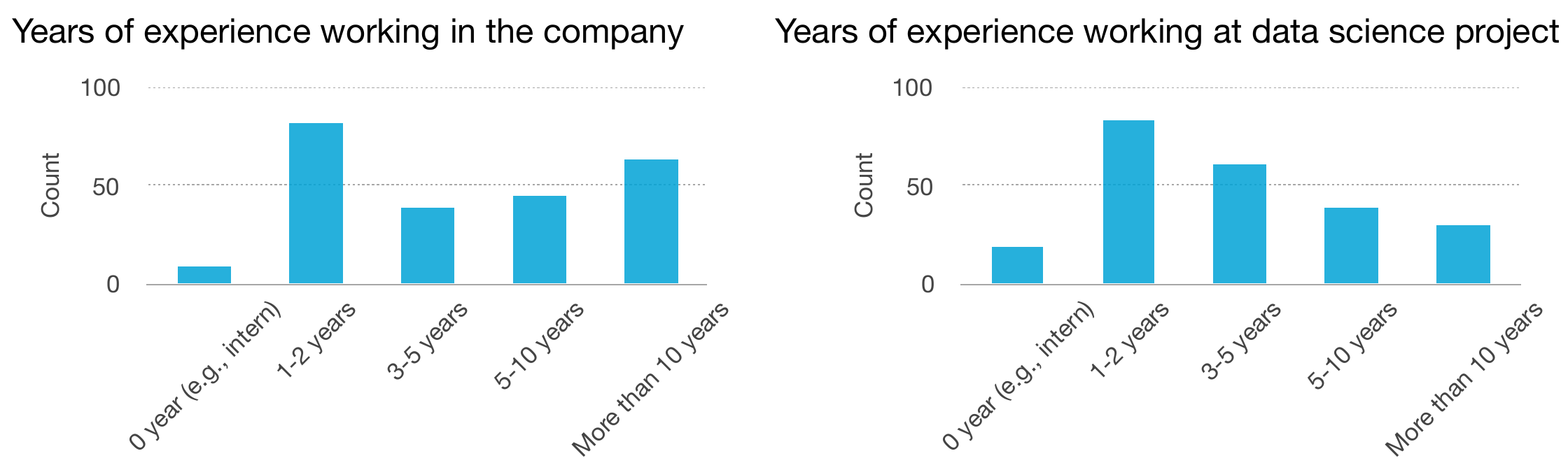}
  \caption{Self-reported background information by respondents. Most respondents have 1-2 years of experience working as a professional, and in DS/ML projects.}
  \label{fig:participants-years}
\end{figure}

We estimate that the Slack channels were read by approximately 2,000 employees in the seven day deployment. Thus, the 239 people who provided data constituted a 11\% participation rate. 
Participants had the option to complete the survey anonymously. Therefore, our knowledge of the participants is derived from their responses to survey items about their roles and their experience on DS/ML projects (Figure~\ref{fig:participants-years}). 

\subsection{Survey Distribution}
We used SurveyGizmo.com to deploy the survey. We posted the survey link and a call for participate in relevant ITCorp's internal Slack channels during May 2020. We wrote one reminder post in the posted channels. Each of the co-authors of this paper also sent the survey link to teammates and friends in the company for a snowball sampling. We collected the last response on May 27, 2020.

\subsection{Expert Interview and Pilot Study}
Wang et al. \cite{wang2019atmseer} introduced an \textit{expert interview} method, where they gathered feedback from two AutoML experts in weekly meetings for two months, while they were building the prototype system. The motivation is that AutoML is a relatively new area of work, so HCI researchers need to leverage AutoML researchers to validate the design prototypes. Following their research, we also hosted weekly meetings with two AutoML experts for two months, while we were developing the survey. In addition, we sent copies of the survey to six external AutoML experts for feedback.

Upon drafting the final version of the survey, we conducted two pilot study sessions -- one with an expert data scientist, and the other one with a citizen data scientist (machine learning engineer). Through the pilot study, we measured the time for participants to finish the survey, adjusted the survey layout, corrected typos, and replaced confusing phrases. Depending on how many lifecycle stages a participant was involved in, the number of questions needed to answer may vary. Thus, the survey could take 5 to 30 minutes for different participants. 

\section{Result}
    
All our participants are employees from ITCorp, a multinational IT company with various departments and thousands of employees. The survey was taken by 239 participants (23 females, 40 males, 4 other, rest undisclosed), among which 22 disqualified were removed from further analysis (e.g., they indicated they had no experience in DS/ML project). All qualified participants had experience working in data science projects (N=217), and 69 of them had more than 5 years of experience, 61 had 3-5 years of experience, and the rest less than 3 years. 

Among the 217 qualified participants, 75 completed the full survey, while the rest partially completed the survey, but failed to complete some part of the survey that asked detailed questions about each stage of the DS/ML lifecycle. In the following sections, we report all data that were collected, including those from participants who partially completed.

Next, we will report and discuss survey results in three subsections: 1) we discuss respondents' primary role in their DS/ML projects and the stages of the projects that they are involved in; 2) we discuss the current degrees of automation that respondents reported as well as their preferred automation levels for each stage; 3) we discuss the information and explanation that the respondents think they need at each stage.

\subsection{Roles and Data Science Life Cycle}\label{result-rolesandDS}
\begin{figure}
      \centering
      \includegraphics{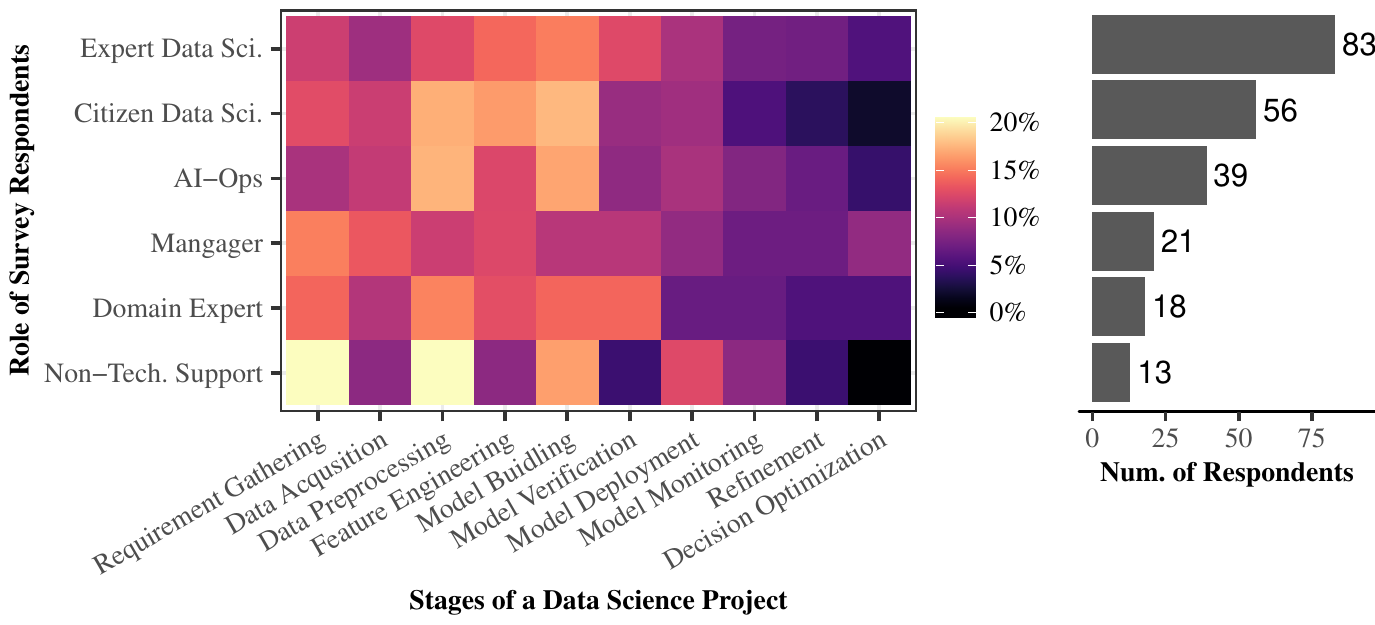}
      \caption{Participation rate of different roles across different stages in a data science project. Left: Different roles participate in different stages at a varying level. Right: The number of respondents corresponding to each role of the left graph.}
      \label{fig:participants-stages}
    \end{figure}

Practitioners of different roles do not participate equally in different stages of the DS/ML lifecycle (Fig.~\ref{fig:participants-stages}
). For example, \textit{expert data scientists} (first row in Fig.~\ref{fig:participants-stages}) reported participating in all the stages of the life cycle, and they are most needed in the technical, middle stages: \textit{data preprocessing, feature engineering, modeling building, model verification, and model deployment}. \textit{Citizen data scientists} are very similar to \textit{expert data scientists}, but they are less likely involved in the \textit{model refinement} or \textit{decision optimization} stage, because they often build models for short-term projects or for learning purposes, which do not have the full lifecycle of using the model in \textit{decision making and optimization}.

By contrast, \textit{domain experts} and\textit{non-technical supporting} roles mostly work in the early stages, such as the \textit{requirement gathering} stage and the \textit{data pre-processing} stages, as these stages are often the places that they contribute human expertise into the DS/ML models, through collaborating with data scientists. \textit{AI/ML Ops} are engineers supporting data scientists, and sometimes also need to implement a model and deploy it as a scalable service, (Fig.~\ref{fig:participants-stages} reflects this practice). \textit{Managers and leaders} are also following through the whole process.

One thing worth mentioning is that the heat map in Fig.~\ref{fig:participants-stages} is normalized on each row, but not on the column dimension. That suggests that we should not compare the absolute percentage across different rows in one column. For example, this chart should not be interpreted as more \textit{non-technical supporting }roles participate or contribute to \textit{requirement gathering} stage than the other roles. 

In Fig.~\ref{fig:participants-stages} to the right, there are the number of respondents who self-identified their particular role in a DS/ML project. In our survey, we also have an ``other'' option, and 16 respondents selected that option with an open-ended answer (e.g., ``data miner'' and ``data engineer''). We coded those answers into the 6 categories of roles (e.g., ``data engineer'' as AI/ML Ops). From these numbers, we can see the majority of survey respondents are technical DS/ML practitioners (e.g., data scientists and AI/ML Ops), which also reflects the norm of the private sector's data science team formation. 

\subsection{Data Science Roles, Their Current and Preferred Levels of Automation, and Different Stages of Lifecycle}
In this subsection, we take a closer look at data science workers' current 
level of automation, and what their preferred level of automation would be in the future. The current and preferred level of automation are also associated with different stages of the DS lifecycle.

\begin{figure}
  \centering
  \includegraphics{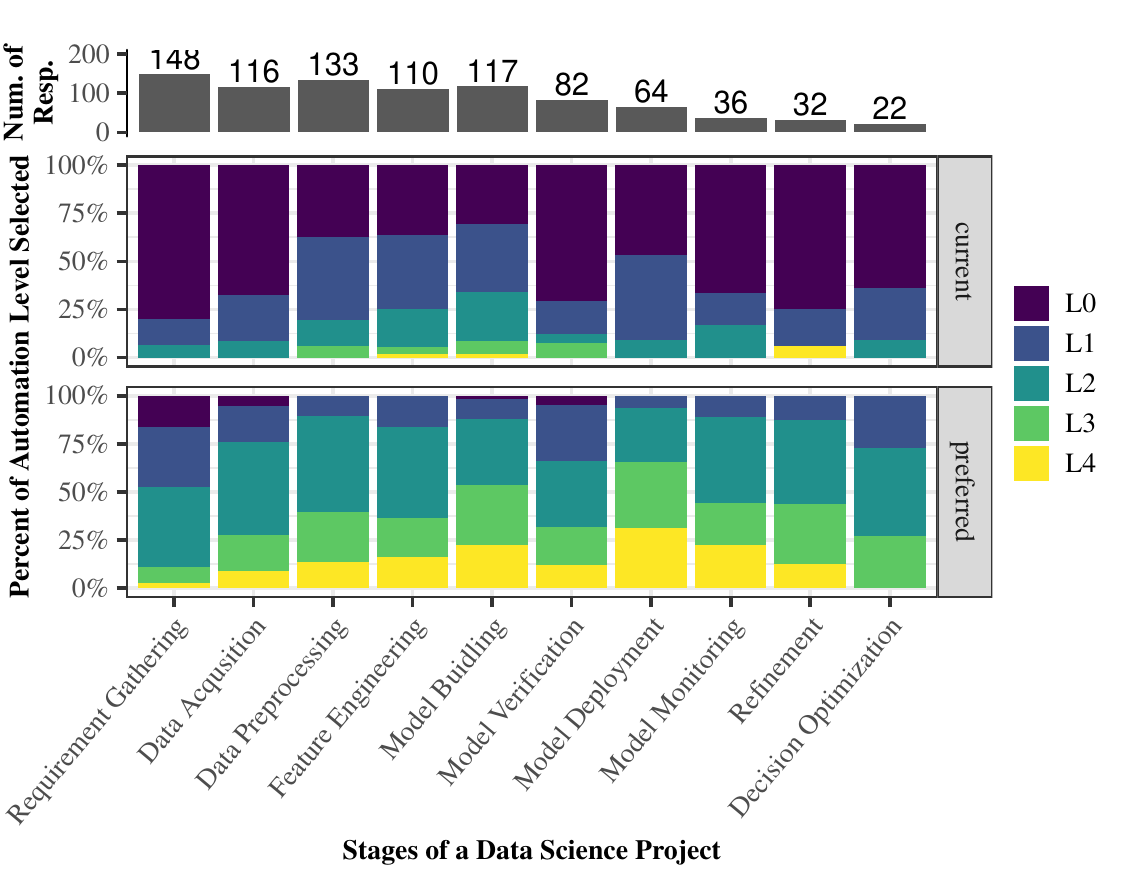}
  \caption{Comparison of respondents' current practice's level of automation and their desired levels of automation in each stage. The small bar chart at the top shows the number of responses collected for each stage.}
  \label{fig:automation-ideal-current}
\end{figure}

When taking all the respondents as a whole, it is observed in  Fig. ~\ref{fig:automation-ideal-current} that there is a clear gap between the levels of automation in their current DS work practices, and preferred automation in the future. The majority of respondents reported that their current work is at automation L0, which is ``No automation, human performs the task''. Some participants reported L1 or even L2 levels of automation (i.e., ``human-directed automation'' and ``system-suggested human-directed'' respectively) in their current work practice, and these automation activities happened often in the more technical stages of the DS lifecycle (e.g., \textit{data pre-processing, feature engineering, modeling building, and model deployment}). These findings echo the existing trend that \autoai{} system development and algorithm research work focus much more on the technical stages of the lifecycle~\cite{zoller2019survey,khurana2016cognito}. However, these degrees of automation are far less than what the respondents desired: participants reported that they prefer at least L1 automation across all stages, with the only exceptions in \textit{requirement gathering} and \textit{model verification} where a number of participants still prefer L0. The median across all the stages is L2 -- human-guided automation. 

In some of the stages, when asked about future preferences, a few respondents indicated that they want full automation (L4) over other automation levels. The \textit{Model deployment} stage had the highest full automation preference, but it was still not the top choice of people. On average across stages, full automation was only preferred by 14\% of respondents. Human-directed automation (L2) was preferred by most respondents (42\%), while system-directed automation (L3) was the second preference (22\%). This suggests that users of \autoai{} would always like to be informed and have the control with the system to some degree. A full end-to-end automated DS lifecycle was not what people wanted. End-to-end \autoai{} systems should always have human-in-the-loop.

There seems to be a trend in the results of the preferred levels of automation: in general, the desired levels of automation increases along with the lifecycle stages moving from less technical ones (e.g., \textit{requirement gathering}) into the more technical ones (e.g., \textit{model building}). L2 (System-suggested human directed automation), L3 (system-directed automation), and L4 (full automation) are the levels when the human shifts some control and decision power to the system, and the \autoai{} system starts to take agency. And L2, L3, and L4 together took the majority vote of each stages. 

In summary, these results suggest that people definitely welcome more automation to help with their DS/ML projects, and there is a huge gap between what they use today and what they want tomorrow. However, people also do not want \textbf{over-automated systems} in the human-centered tasks (i.e., \textit{requirement gathering, model verification, and decision optimization}).

\begin{figure}
  \centering
  \includegraphics[width=\textwidth]{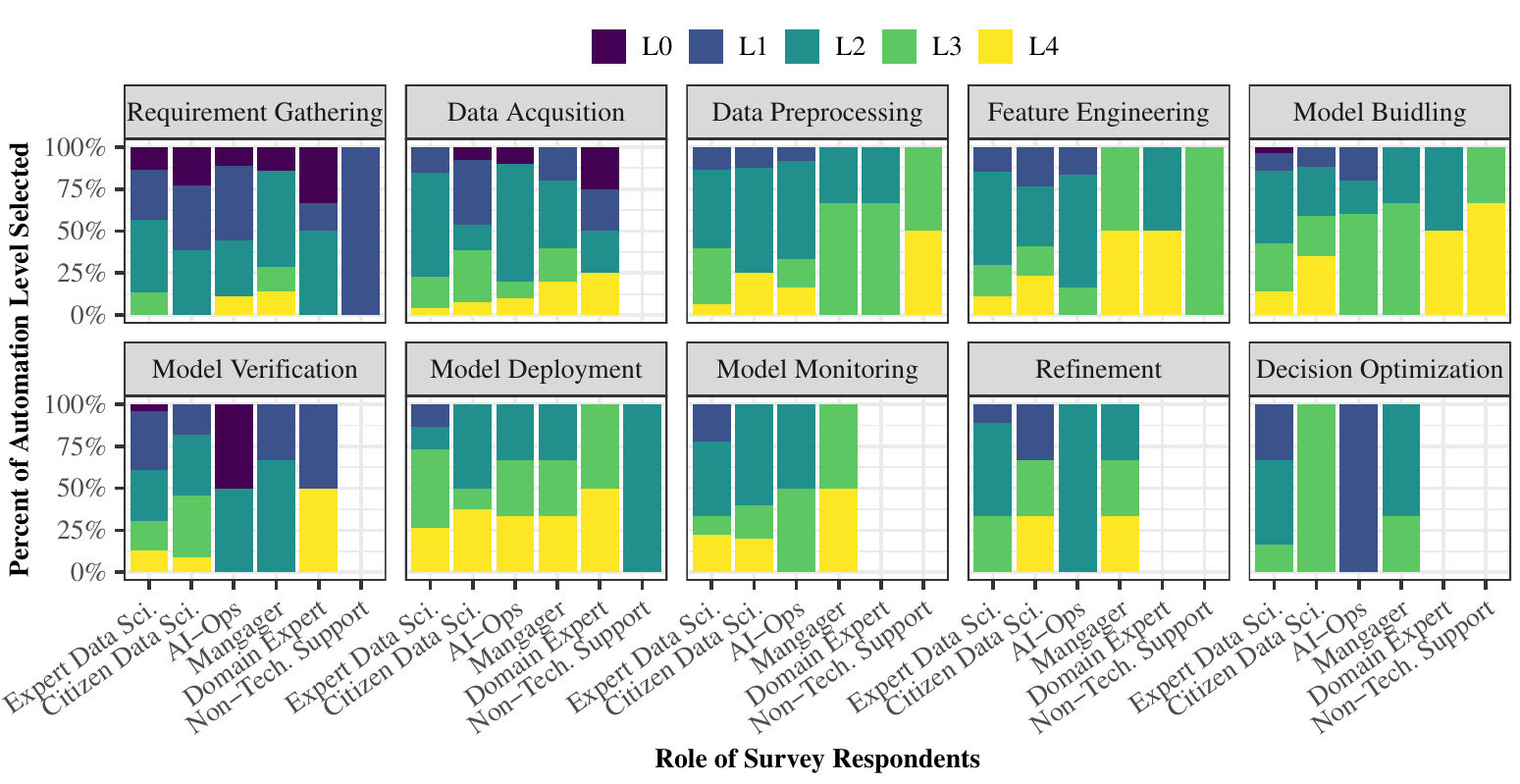}
  \caption{Preferences for automation of different roles for various stages. Blank columns indicate that we did not collect any data for that practitioner role at that  stage.}
  \label{fig:role-stage-prefer-automation}
\end{figure}

To have an in-depth examination of people's preferred levels of automation, we plotted  Fig. ~\ref{fig:role-stage-prefer-automation} with finer-grained level of automation preferences across different stages, and across different roles. In the figure, we have 10 charts, with each representing one stage. Within the same stage, people's preferred levels of automation differ greatly depending on their roles and amount of involvement in each stage, and how familiar they are with that task. For example, non-technically savvy roles (e.g., \textit{non-technical supporter} and\textit{ domain experts}), did not participate in some stages at all (e.g. \textit{model building} or \textit{model refinement}); and for most stages (except \textit{requirement gathering}), they prefered high levels of automation. It is worth noting that participants across all roles agreed that \textit{requirement gathering} should remain a relatively manual process.

Data scientists, both expert and citizen scientists, tend to be cautious about automation. Only a few of them expressed interest in fully automating (L4) \textit{feature engineering, model building, model verification, and runtime monitoring}. For example, in \textit{model verification}, they prefer system-suggested and human-directed (L3) and system-directed automation (L2) over too little automation (L1/L0) or too much automation (L4). 

AI-Ops had a more conservative perspective toward automation than other roles, they only have a majority preference of full automation (L4) in the \textit{model deployment} stage, some on \textit{data acquisition} and \textit{data pre-processing}, but for the rest stages, they would strongly prefer to have human involvement. 

Above all, there is a clear consensus among different roles that \textit{model deployment, feature engineering, and model building} are the places where practitioners want higher levels of automation. This suggests an opportunity for researchers and system builders to prioritize automation work on these stages. On the other hand, all roles agree that less automation is desired in \textit{requirements gathering and decision optimization} stages, this may be due to the fact these stages are currently labor-intensive human efforts, and it is difficult for our participants to even imagine how the automation would look like in these stages in the future.

\begin{figure}
  \centering
  \includegraphics{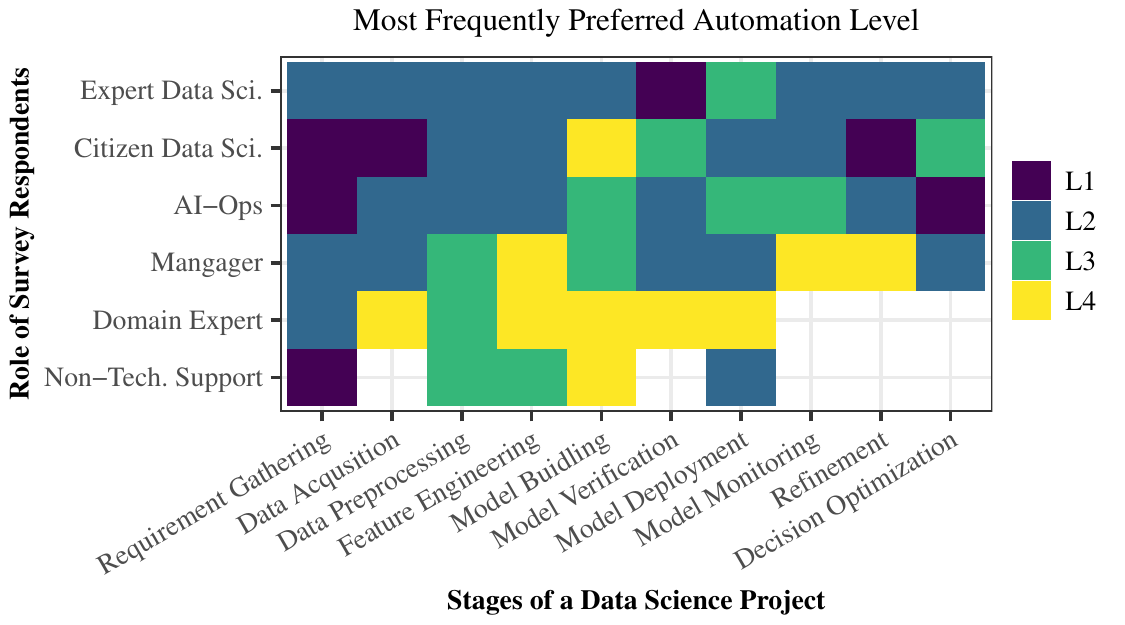}
  \caption{Most preferred level of automation by each role in different stages. Blank cells indicate no data collected for that particular role and stage.}
  \label{fig:role-prefer-automation}
\end{figure}

Fig. ~\ref{fig:role-prefer-automation} shows another view of the top preference of the level of automation for different roles across different stages. Echoing our observation from Fig. ~\ref{fig:role-stage-prefer-automation}, the top preferences for \textit{data scientists} are mostly \textit{system-suggested human-directed automation (L2)} and \textit{system--directed automation (L3)}. However, \textit{domain experts} and \textit{non-technical supporters} generally desire a higher level of automation across different stages such as \textit{full automation (L4)} or at least \textit{system--directed automation (L3)}.

\subsection{XAI in \autoai}
\begin{figure}
      \centering
      \includegraphics{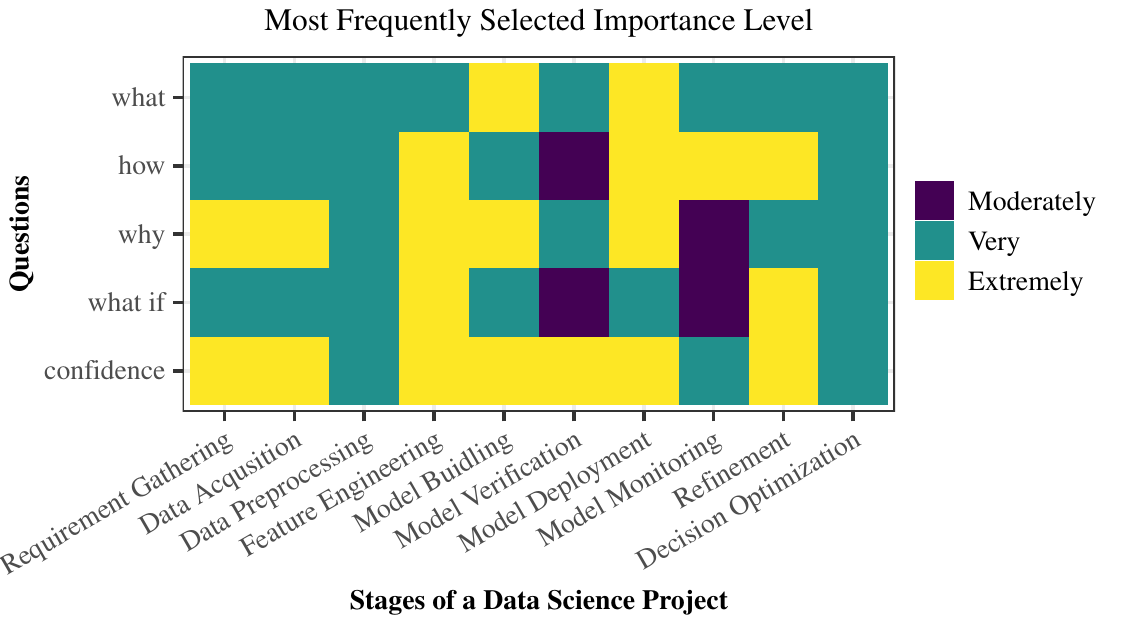}
      \caption{Most frequently chosen importance level for each category of explainability.}
      \Description{}

   \label{fig:explanation}
    \end{figure}

Participants indicated their needs for explainability for each stages of DS lifecycle, if automated, by rating the importance of five prototypical questions, as described in Section~\ref{survey_design}. Figure~\ref{fig:explanation} presents the mode---the most frequently selected importance level of each of the question categories for each DS lifecycle stage. For three of the stages where the explainability question was asked for each sub-task, we use the mode of a participant's ratings for all sub-tasks within a stage as the participant's response for the stage. Results demonstrate that explainability is an instrumental part of the \autoai{} user experience, indicated by the fact that the mode of ratings was at least \textit{Moderately important} for all questions in all stages. We make the following observations based on Figure~\ref{fig:explanation}.

\textit{Confidence} information was rated to have the highest importance, being \textit{Extremely important} for the majority of the stages. It suggests a primary user need of \autoai{} is to understand the AI's certainty to either feel more confident about the automation or to exercise more caution.
    
\textit{How} and \textit{Why} questions were rated to be \textit{Very important} to \textit{Extremely important}, just following \textit{Certainty} information. They show that people want a level of understanding on the automation's working mechanism for almost all stages. Such understanding is both at a global level on how the automation works, and at individual level on the rationale for a specific execution.
    
\textit{What} questions, to understand descriptively the scope of automation output or functions, were considered \textit{Extremely important} for Model Building and Model Deployment. The two stages also have a relatively large number and wide scope of sub-tasks. It suggests that there are rich activities involved in model building and deployment, and users may need help understanding what activities are covered by the automation. 
    
\textit{What-if} questions, asking about counterfactual outcomes, were considered extremely important for Feature Engineering and Model Refinement, suggesting that for these stages people are interested in exploring the decision space---not just accepting the automation's single recommendations.
    
 Across stages, people wanted the most explainability in Feature Engineering and Model Deployment. Interestingly, these are also the stages in which people wanted the highest level of automation. It suggests that there might be increasing needs for explainability with increasing levels of automation. People wanted the least explainability in model verification and model monitoring. There may be two reasons. One is that both stages are concerned with more mechanical tasks such as quantifying model behaviors and generating reports. These activities are relatively straightforward to understand. Second, people also desired less automation and more human autonomy in these stages, so explainability of automation might be less of a concern.

\begin{figure}
      \centering
      \includegraphics{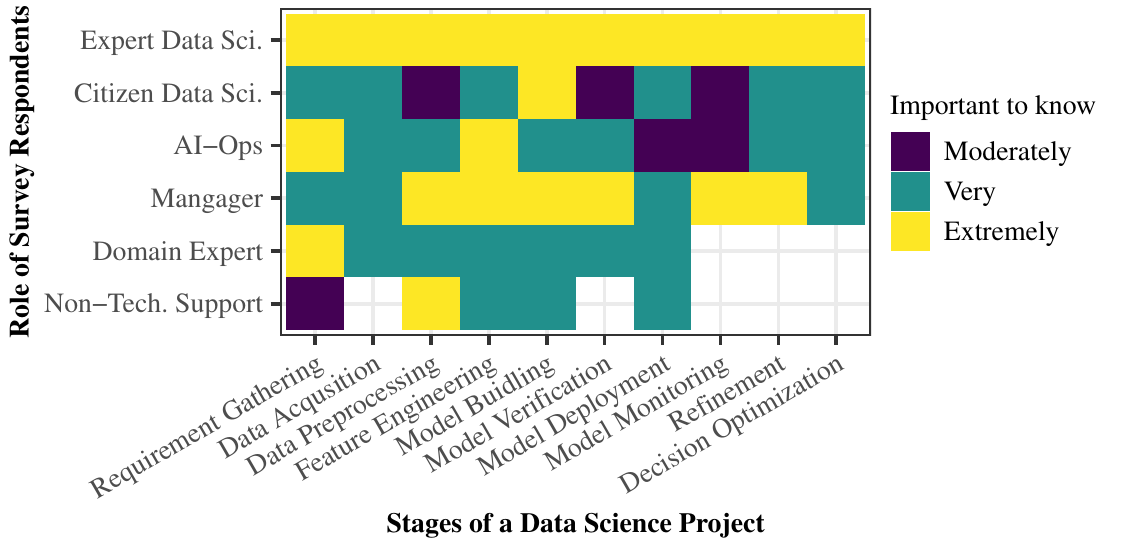}
      \caption{Most frequently selected importance level for explainability questions by different roles across stages.}
      \Description{}

   \label{fig:role-exp}
    \end{figure}
    
We then plot the distribution of importance ratings for explainability questions by different roles across stages, as shown in Figure~\ref{fig:role-exp}. The patterns suggest that expert data scientists have the highest needs for explainability, with the mode of explainability ratings in all stages being \textit{Extremely important}. Managers also had strong needs for explainability in all technical stages of model development and refinement. Interestingly, citizen data scientists had the lowest needs for explainability of all roles, possibly because the nature of their involvement with DS/ML projects is not always tied to critical business outcomes.  

\section{Discussion}
General AI systems, which can perform intellectual tasks on a par with humans, have long been envisioned by humanity, since well before Turing's time~\cite{saygin2000turing}. This vision drives many researchers to pursue fully automated machine learning and data science as the ultimate research goal. However, there are various unresolved technical problems along each stage of the data science and machine learning lifecycle, not even to mention the human-AI interaction challenges on top of those. This ultimate goal will remain elusive for years --- if not decades --- to come.

Hence, it is even more important to determine the appropriate level of automation at the right moment, and for the right user. Knowing these role-specific levels can guide us to design a more desirable and more user-friendly AutoDS system to help diverse types of users to get their work done more effectively. For this purpose, we conducted the first large-scale user study to solicit data science and machine learning practitioners' feedback from professionals who work in DS/ML projects today. In this section, we organize the discussion in four subsections, we start with design implications for \textbf{human-in-the-loop} \autoai{} systems; then, we specifically focus on the designs for non-technical DS/ML practitioners (e.g., domain experts, communicators), as their needs are often overlooked in today's \autoai{} system design; we then move on to the explainable and trust AI issues in \autoai{}, as it is fairly a new research topic; lastly, we end this section with limitations and future research directions.

\subsection{Human-Centered Design for \autoai}
Our results show that for different stages in the DS/ML lifecycle, different levels of automation are desired by different user roles. For example, \textit{domain experts} prefer L4 level of full automation in \textit{model building} stage; whereas \textit{AI/ML Ops engineers} prefer L3 (system-guided automation). This result suggests ``one-solution-fits-all'' \autoai{} simply does not exist. Instead, a human-in-the-loop \autoai{} system will need to be able to meet the requirements from different personas. In the same example, \textit{domain experts} do not understand the \textit{model building} details, and they do not need to understand those details. Thus for them, there are no controllability needs; they may be satisfied to interact with the L4 \autoai{} system in \textit{model building} as an opaquebox, as long as they can get their desired explanation and transparency information. But for the same \textit{domain expert}, they would be challenged if the \autoai{} system is still at L4 for \textit{requirement gathering} stage. That is where they express their domain-specific requirements and contribute their human knowledge and expertise for the entire project. An L2 level (human-guided automation) can give the control and agency back to humans, and that is what domain experts would appreciate. 

Most of respondents reported that they preferred L2 and L3 levels of automation in most stages. Thus, a human-guided automation or a system-suggestion with human-guided \autoai{} will be dominant for the near future. Researchers can shift some of their focus from aiming for a fully automated \autoai{} system, to resolving some of the immediate user needs. 

There are still plenty of research opportunities for this near-term research goal. For example, all user personas claimed that \textit{feature engineering} was one important stage that they participated in (Fig.~\ref{fig:participants-stages}), but their desired automation levels were different. Most data scientists preferred L2 and L3, as they often lacked the necessary domain knowledge when generating new features and deciding on the generated features. Without domain knowledge to guide the feature generation, sometimes there may be meaningless new features (e.g., absolute value of gender as a new feature) that lead to a seemingly good model result (i.e., over fitting). 

One good example along this direction is the KAFE project for automated feature engineering~\cite{kafe}, where the algorithm can automatically search Wikipedia pages for needed feature information (e.g., norms for human body weight) using the column characteristics (e.g., with keyword ``body weight'') to gather domain knowledge about the data or features. Based on the automatically-acquired information, it can \textbf{guide} novel feature generation (e.g., body mass index (BMI)) leveraging the domain knowledge that it has collected, and data scientist users can not only decide whether to apply the feature information or not, but can also obtain an interpretable trace on how the new feature was derived. If this type of algorithms can be incorporated into an \autoai{} system, it will provide the desired L3 (system-guided) level of automation for \textit{data scientists}. We can find many of these example algorithms and design suggestions for each cell of Fig.~\ref{fig:role-prefer-automation}, such as OneButtonMachine algorithm~\cite{lam2017one} for human-guided automation (L2) for \textit{data scientists} during \textit{data acquisition and fusion}; FactSheets~\cite{arnold2019factsheets} or ModelCard~\cite{mitchell2019model} for AI/ML Ops during \textit{model verification} stage, and similarly for other automation approaches to selected portions of the data science lifecycle. 

Last but not least, we should always remember that data science and machine learning are not single user jobs, as suggested in our Result~\ref{result-rolesandDS} and by previous literature~\cite{amyzhang2020,clarke2019training, muller2019datascience, saltz2018exploring}. Thus, an \autoai{} system should be able to support multi-user communications and collaborations in every stage, as well as across stages. The minimum version control and action logging function should be supported so that a \textit{data scientist} in \textit{data acquisition} stage knows what the \autoai{} and the \textit{domain expert} did in the previous \textit{requirement gathering and problem formulation} stage, and why they did it.

\subsection{\autoai{} for Non-Technical DS/ML Practitioners}
``Democratizing AI'' refers to the effort of lowering the boundaries for non-technical users. Unlike data scientists or software engineers,  who either got formal training of DS/ML knowledge or have a capability to quickly catch up DS/ML programming skills~\cite{kim2016emerging}, non-technical practitioners often lack the foundation or motivation for systematically learning DS/ML skills. They contribute to a DS/ML project with their domain knowledge and human expertise in the problem context. Thus, many companies and research groups share this ``democratizing AI'' vision, and they have built various end-customer-facing user interfaces~\cite{cai2019human} or higher abstraction level programming packages~\cite{gao2018panda} to fulfill this goal. Yet, in reality, all current \autoai{} systems (e.g., Google AutoML, H2O, DataRobot) focus heavily on the technical personas (e.g., data scientists and AI/ML Ops engineers). 

It is an utterly different challenge for \autoai{} design when targeting non-technical users (instead of data scientist users) for democratizing AI. As shown in Fig.~\ref{fig:role-prefer-automation}, the starting and ending stages (e.g., requirement gathering and decision optimization) are expected to have more human judgment and less automation. These stages are also the places where non-technical users take the most responsibility (Fig. ~\ref{fig:participants-stages}. Unsurprisingly, the automation capabilities for those stages currently are extremely limited --- the more human knowledge is needed, the harder it is to design an automation for integrating such human knowledge. This limited automation presents a great opportunity for \autoai{} researchers and system designers. 
Even more or less limited functionality like automated report generation~\cite{automaticst}, form-based requirement gathering~\cite{wang2020autoai}, or enforcing stakeholder constraints (e.g., fairness threshold)~\cite{liu2019admm} can enable non-technical users to build, steer, and understand the AI solution to some extent. 
In addition, recent advances in machine teaching hint at a new way of automatically integrating domain knowledge from subject matter experts~\cite{Wilder2020complement}.
The previous example of domain-knowledge assisted feature engineering (KAFE) ~\cite{kafe} can also use some help from non-technical users, so the automated knowledge solicitation can be more effectively. 
With these add-on features on \autoai{}, a non-technical practitioner in DS/ML projects can directly and more flexibly input their feedback, and influence \autoai{} behaviors and model outcomes without technical experts' help. 

We speculate that future \autoai{} systems are likely to be more and more powerful, and the distinction between technical citizen data scientists and non-technical practitioners may become blurry. More and more colleges may include \autoai{} tools in their curriculum, as they did with SPSS, R, and SAS today. So a social science student and an epidemiologist is less likely to need a statistician to run every analysis, because they can do those simple tasks themselves. However, the data scientists job will not go away. When people encounter challenging DS/ML problems, they will still need to find professional data scientists.

\subsection{Explainable and Trustworthy \autoai}

Trust plays a vital role in users' acceptance and adoption of new technologies. In order to foster trust in \autoai{} systems, we believe it is imperative to allow users to \textit{understand} the automation by making the \autoai{} system explainable~\cite{JaimieDrozdal2020}. For example, data scientists may demand an explanation for every decision the \autoai{} system makes, so they can decide if they need to make adjustments, even at the source code level~\cite{karl2020}. Managers may demand an explanation for the features and model chosen by \autoai{}, so they could be assured that the model complies with regulatory and client requirements. Domain experts may demand an explanation of what domain knowledge the \autoai{} systems has access to, to make sure all grounds are covered. Prior work also suggests that explanations could help set appropriate expectations for \textit{imperfect} automation and could make people more forgiving towards system's mistakes~\cite{kocielnik2019will,ashktorab2019resilient}. For example, understanding the scope of automated feature engineering that an \autoai{} system is able to perform, could prepare the users for certain limitations. Acknowledging low confidence for a particular operation could alert users to intervene early and avoid adverse outcomes. 

It is important to recognize that different types of explanation are requested in the above scenarios. Considering the complex set of technologies involved in \autoai{} systems, it is infeasible and also undesirable to expose all its inner working logic in all stages to all users. Our work provides a framework of explainability types for \autoai{} systems and guidance for prioritizing different types of explanation. Specifically, based on prior work~\cite{lim2009and,lim2009assessing,liao2020questioning,rader2018explanations}, we suggest providing explanations to answer five types of user questions: \textit{What},\textit{How}, \textit{Why},\textit{What-if} and \textit{Confidence}. They are applicable to all stages of ML/DS lifecycle but some categories, including \textit{Confidence}, \textit{How} and \textit{Why} should be prioritized in general. Meanwhile the prioritization of explainability needs and requirement to address them depend highly on the user group and the ML tasks, calling for empirical, human-centered approaches to design explanation features for \autoai{} systems. 

Our work only begins to consider high-level user needs for explainable \autoai. Future work should explore viable solutions to generate explanations that address these questions, both on how to generate the algorithmic explanations and how to present them to users. It would also be advisable to consider the \textit{cost} of generating different types of explanations. For example, explanations for \textit{What} and \textit{How} questions could be easier to realize by summarizing the output and process of implemented \autoai{} solutions, while \textit{Why}, \textit{What-if} and \textit{Confidence} could be challenging to support depending on the underlying technologies, and may require developing a separate set of techniques to address. 

In implementing the user interface for rendering these explanation information, we can use some help from the information visualization community. Recent papers experimented with various new visualization implementations~\cite{karl2020,wang2019atmseer} to surface different types of information from \autoai{} systems, yet we have not seen any of these projects go beyond a system prototype. We look forward to more research teams joining this effort of designing human-in-the-loop explainable \autoai{}. 

\subsection{Limitations and Future Research Directions}

While this study provided valuable insights into user needs for \autoai{} with a large scale survey study, more work needs to be performed to obtain a holistic picture. For instance, our participants sample is limited to ITCorp. Although they span a wide range of product groups and geographical locations, it is possible that practices of DS/ML project are different in other organizations. It would be desirable to replicate the results with a larger and more diverse pool of participants from different organizations. Moreover, the current work only inquired into what people do, what people prefer, and what people need. We plan to conduct additional studies in the future to further understand the \textit{reasons why} people preferred certain automation levels and types of explanations.

Our introduced framework can also be used to analyze and evaluate \autoai{} systems in terms of end-to-end coverage, level of automation, and supported human-in-the-loop capabilities, thus further helping system designers and builders to understand if a particular approach satisfies their requirements.
The gaps between current and future preferred levels of automation exposed in this study also help with steering the research direction of automated systems for Data Science. There are clear trends for each stage, and this can at least serve as a ranking of the various opportunities for automation.

When the interaction with the user is facilitated not only to satisfy the requirements of the human, but also as valuable machine-understandable information for the AI system itself, the AI system may be able to learn from the history of user interactions, and may gradually improve its capability to bootstrap even more sophisticated automations. 

Last but not least, \autoai{} presents multiple rapidly evolving research topics. The findings from our survey study may change along with the widely adoption of the \autoai{} technology. Nonetheless, we believe our proposed human-in-the-loop \autoai{} framework will be capable of evolving along with new findings and methods, and will help with the evolving research endeavor as well as the soon-to-come user adoption.

\section{Conclusion}
In this paper, we proposed a human-in-the-loop \autoai{} framework with four dimensions: roles, stages, levels of automation, and types of explanation. Then, we used the framework to design a large scale online survey to gather usage perspectives from data science and machine learning practitioners. We found an exciting opportunity for \autoai{} research and system development, because there is a huge gap between the automation level in people's current work practice, and the future automation level that they prefer. However, such research and development effort should be more directed to meet various user personas' specific needs. The level of automation and the type of explanation can vary a lot depending on who the user is, which lifecycle stage the user works in, and what the task is. We argue there is no need of a fully automated data science and machine learning work, but rather that a human-in-the-loop explainable \autoai{} is a promising future. We call for fellow researchers to join this important HCI-AI collaborative research endeavor.

\begin{acks}
BLANK.
\end{acks}

\bibliographystyle{ACM-Reference-Format}
\bibliography{main}

\newpage

\appendix

\section{Human-in-the-Loop AutoML Framework}
\subsection{Roles} \label{appendix:role}
\begin{itemize}
  \item Expert Data Scientist / Researcher (who build new model algorithm and engage in all stages of data science lifecycle)
  \item Citizen Data Scientist (who occasionally build models or engage in ML projects, e.g., some ML engineer, business analyst, data volunteer)
  \item AI-Ops / ML-Ops (who prepare data or integrate model into application, e.g., data engineer, some ML engineer, application developer)
  \item Lead/Manager
  \item Domain Expert / Subject Matter Expert
  \item Non-Technical Supporting Roles (who support various stages of the data science lifecycle e.g., Design, Comm, Sales, Consultant, Data Labeler)

\end{itemize}

\subsection{DS/ML Lifecycle with 10 Stages and 43 Sub-Tasks} \label{appendix:lifecycle}
\begin{itemize}
   \item 1. Requirement Gathering and Problem Formulation (e.g, System can gather domain knowledge or regulation requirements and define a data science problem.)
       \begin{itemize}
        \item 1.1 Requirement Collection (requirement Gathering)
        \item 1.2. Domain knowledge specification (system automatically finds relevant sources of domain knowledge, such as excerpts from books,, wikipedia, kaggle, etc.)
        \item 1.3. Regulation gathering (system automatically fetches the relevant regulation )
        \item 1.4 Problem formulation (system define the type of classification and computation objective)
    \end{itemize}
   
    \item 2. Data Acquisition and Governance (e.g, System can find relevant dataset and combine datasets for given a data science problem.)
        \begin{itemize}
        \item  2.1 Data Acquisition (system can find a list of datasets and meta-data suitable for the given problem)
        \item 2.2 Data Fusion (system can find identical features across datasets and join datasets by these features)
        \item 2.3 Data Governance and Security (System can enforce regulations, fairness, and rules on models and data in form of constraints)

    \end{itemize}
    
    \item 3. Data Readiness, Preparation, and Cleaning (e.g, System can assess data quality, detect data noice, and clean the data.)
        \begin{itemize}
        \item 3.1 Labeling and Validating Label  (automatic clustering, automatically assign labels to some instances)
        \item 3.2 Data Profiling and Exploratory Data Analysis   (Automatically assess overall data quality, detect noise in labels, skewness and correlations, class imbalance, and target distributions).
        \item 3.3 Data Cleaning and Quality Remediation (auto Cleaning or filtering with explanations).
        \item 3.4 Data Bias Detection and Mitigation  (e.g., find bias, homogeneity in target labels and propose de-bias techniques).
    \end{itemize}
    
    \item 4. Feature Engineering (e.g, System can apply various feature transformation and encoding functions, and select the best subset of features).
        \begin{itemize}
        \item 4.1 Variable Encoding (one hot encoding, ordinal encoding, system can try to determine the best encoding strategy for certain features).
        \item 4.2 Feature Transformation (eg Principal Component Decomposition/PCD, Arithmetic, Trigonometric functions, system can determine a ranked list of features that should be added or dropped).
        \item 4.3 Feature Aggregation (eg aggregation based on time or location).
        \item 4.4 Feature Selection (remove unnecessary and repetitive features to create an optimal feature set).
        \item 4.5 Feature Augmentation with Domain Knowledge (eg BMI based on body height and weight) (automatically identify column concepts and link them to derive useful new features based on domain knowledge such as that in Wikipedia).
        \end{itemize}
    
    \item 5. Model Building and Training (e.g, System can split train-test dataset, select the best algorithm or neural architecture, and find an optimized set of hyperparameters).
        \begin{itemize}
        \item 5.1 Sub-Sampling and Train-Test Splitting (automatically determine the best strategy for train-test splitting and sub-sampling methodology for large data).
        \item 5.2 Algorithm and Neural Architecture Selection (System offers methods for estimator searching and optimization).
        \item 5.3 Model parameter tuning (System offers a variety of hyper-parameter optimization methods).
        \item 5.4 Model Ensemble (automatically find a good ensemble that improves performance over the best single estimator).
        \item 5.5 Benchmark Testing (System can test new model with benchmark data).
        \item 5.6 Model Evaluation ( System can compare models with traditional ML metrics and with business and Domain-specific KPI).
        \end{itemize}

    \item 6. Model Presentation and Verification (e.g, System can prepare reports and dashboard for a model, and verify model with stakeholders.)
        \begin{itemize}
        \item 6.1 Preparing Reports (System can automatically generate reports in Word, HTML and slides).
        \item 6.2 Verifying Model (System can automatically verify the model with stakeholders).
        \item 6.3 Stakeholder Buy-in (System can suggest, confirm, and track stakeholder buy in the model).
        \item 6.4 Detecting and Mitigating Model Bias  (System can detect and mitigate model bias).

        \end{itemize}
    
    \item 7. Model Deployment (e.g, System can dockerize model as microservice, deploy, and conduct A/B test with different versions of the model.)
        \begin{itemize}
        \item   7.1 Dockerize, Scalable Microservice, and Cluster  (system can automatically create scalable microserveices, support batch testing, and set up elastic cloud service).
        \item 7.2 Regulation compliance, Fraud, and Security (system can automatically test regulation compliance and evaluate security risks).
        \item 7.3 Integration into existing software system  (system can automatically integrate the new model into existing software systems or applications).
        \item 7.4 Load-testing and balancing (system can automatically run through various load balancing options and perform peak-testing and report numbers).
        \item 7.5 A/B testing of deploying different models (system can automatically do A/B testing for deployment and replace/rollback model versions based on some criteria).
        \end{itemize}
    
    \item 8. Model Runtime Monitoring (e.g, System can monitor performance drift, security, and fraud risk.)
        \begin{itemize}
        \item 8.1 Dashboard Setup (system can automatically suggest metrics to monitor and deploy instrumentation code, storage and reporting dashboard).
        \item 8.2 Performance Drift Detection (system can automatically detect performance drift in the incoming data distribution and load and alert the administrator periodically or based on a certain threshold).
        \item 8.3 Monitoring A/B testing (system can automatically monitor and visualize A/B testing results different versions of deployed model).
        \item 8.4 Regulation compliance, Fraud, and Security (system can automatically test regulation compliance and evaluate security risks).

    \end{itemize}
    
    \item 9. Post-Deployment Model Refinement (e.g, System can refine the model with deployment data to optimize Key-Performance-Indicator-based (KPI) not only for accuracy metrics. )
    \begin{itemize}
        \item  9.1 Refining Model for new data  (system can automatically use runtime performance log data and perform model retrain and re-deploy).
        \item 9.2 Adding new labels to data or Relabeling (system can automatically labeling new data or changing existing labels).
        \item 9.3 Business Key-Performance-Indicator-based (KPI) model refinement (system can automatically monitor KPIs and retrain model for better KPIs).
    \end{itemize}

    \item 10. Decision Making and Optimization (e.g, System can automatically collect domain knowledge for contextualize decision making and further support decision optimization. )
    \begin{itemize}
        \item  10.1 Domain Knowledge Specification for Decision Optimization (system can automatically collect domain knowledge for contextualize decision making and further support decision optimization).
        \item 10.2 Rendering Dashboard and Visualization ( system can render a visual of the essential trade-offs or other aspects that make up the returned/computed optimal solution).
        \item 10.3 Decision Optimization Forensics (the system can inspect and report the quality of the data lead to this decision).
        \item 10.4 Explaining and Justifying Decision Recommendation (system can generate justification and explanation for the data, the model, and the inference, such as comparing them with the alternative options).
        \item 10.5 Tracking Usage of Recommendations (system can track and self-correct recommendations in response to unsatisfactory user actions).
    \end{itemize}

\end{itemize}

\subsection{Levels of Automation} \label{appendix:level}
\begin{itemize}
    \item \textbf{Level 0 No automation}: No automation, human performs the task.
    \item \textbf{Level 1 Human-directed automation}: System may provide automation functions (e.g., auto-sklearn), human selects methods and provide parameters before executing.
    \item \textbf{Level 2 System-suggested Human-directed}: System decides on parameters/methods/actions, human approves or changes suggested configuration before executing.
    \item \textbf{Level 3 System-directed automation}: System executes automatically and informs human with progress and results, human can intervene anytime.
    \item \textbf{Level 4 Full automation}: System decides everything and executes automatically, human is not in the loop.
\end{itemize}

\subsection{Types of Explanation} \label{appendix:level}
\begin{itemize}
    \item \textbf{What}: What kind of automatic [stage name] can the system do?
    \item \textbf{How}: How does the system perform automatic [stage name]? 
    \item \textbf{Why}: Why did the system execute this way and give this output?  
    \item \textbf{What-if}: Would the system execute differently if...(something changes in the ML task or data)?
    \item \textbf{Confidence}: How confident is the system about this choice of execution? 
\end{itemize}

\end{document}